%% file: main.tex
\RequirePackage[svgnames]{xcolor}

\documentclass[11pt,logo,letterpaper]{berkeley}

\usepackage[utf8]{inputenc} %
\usepackage[T1]{fontenc}    %
\usepackage{lmodern}
\usepackage{textcomp}
\usepackage[svgnames]{xcolor}
\usepackage{asymptote}
\usepackage[comma,authoryear,compress]{natbib}
\usepackage{hyperref}
\usepackage{algorithm}
\usepackage{algorithmicx}
\usepackage{algpseudocode}
\usepackage{microtype}
\usepackage{graphicx}
\expandafter\def\csname ver@subfig.sty\endcsname{}
\usepackage{booktabs} %
\usepackage{float}
\usepackage{bigstrut}
\usepackage{tikz}
\usetikzlibrary{tikzmark}
\makeatletter
\newcommand*\myfontsize{%
  \@setfontsize\myfontsize{7}{8}%
}
\makeatother
\definecolor{myred}{rgb}{0.7, 0.3, 0.0}
\definecolor{myblue}{HTML}{054488}
\definecolor{mygreen}{HTML}{056b34}

\newcolumntype{R}[1]{>{\raggedleft\let\newline\\\arraybackslash\hspace{0pt}}m{#1}}

\usetikzlibrary{intersections}

\definecolor{darkgreen}{rgb}{0.0, 0.42, 0.24}

\usepackage{amsthm}
\usepackage{nicefrac}
\usepackage{subcaption}
\usepackage{caption}
\usepackage{graphicx}
\usepackage{cleveref}
\usepackage{bxcoloremoji}
\usepackage{listings}
\usepackage{float}
\usepackage{longtable}
\lstdefinestyle{python}{
    language=Python,
    basicstyle=\ttfamily\footnotesize,
    keywordstyle=\color{blue}\bfseries,
    commentstyle=\color{green},
    stringstyle=\color{red},
    numberstyle=\tiny\color{gray},
    showstringspaces=false,
    frame=single,
    breaklines=true,
    backgroundcolor=\color{lightgray!20}
}

\usepackage{hyperref}       %
\usepackage{url}            %
\usepackage{booktabs}       %
\usepackage{nicefrac}       %
\usepackage{microtype}      %
\usepackage{graphicx}
\usepackage{subcaption} 
\usepackage{wrapfig}
\usepackage{lipsum}
\usepackage{enumitem}
\usepackage{stackengine}
\usepackage[font=small,labelfont=bf]{caption}
\usepackage{color}
\usepackage{adjustbox}
\usepackage{tikz}
\usepackage{tcolorbox}
\usepackage{rotating}
\usepackage{makecell}
\usepackage{colortbl}    
\usepackage{multirow}    
\usepackage{pifont}      
\usepackage{xcolor}      
\usepackage{array}       
\usepackage{amsmath}
\usepackage{amsfonts}       %
\usepackage{xspace}
\usepackage{mathtools}
\definecolor{oursblue}{RGB}{230,240,255} 

\input{macro}

\newtcolorbox{AIbox}[2][]{aibox,title=#2,#1}
\definecolor{lightblue}{rgb}{0.22,0.45,0.70}%
\definecolor{Gray}{gray}{0.95}
\definecolor{Cornsilk}{rgb}{1.0, 0.97, 0.86}
\usepackage{enumitem}

\makeatother

\definecolor{myred}{rgb}{0.7, 0.3, 0.0}
\definecolor{myblue}{HTML}{054488}
\definecolor{mygreen}{HTML}{056b34}
\definecolor{myorange}{HTML}{ff8800}
\definecolor{mypurple}{HTML}{8400ff}
\definecolor{mypink}{HTML}{f7acb9}

\definecolor{myred}{rgb}{0.7, 0.3, 0.0}
\definecolor{myblue}{HTML}{054488}
\definecolor{mygreen}{HTML}{056b34}
\definecolor{tiktokpink}{HTML}{E91E63}
\definecolor{tiktokpurple}{HTML}{673AB7}
\definecolor{tiktokgray}{HTML}{9E9E9E}

\newcommand{\mytitle}{Integrating Explainable Machine Learning and Mixed-Integer Optimization for Personalized Sleep Quality Intervention}

\usepackage{amssymb}

\title{\mytitle}

\author{
Mahfuz Ahmed Anik$^{1}$, 
Mohsin Mahmud Topu$^{1}$,
Azmine Toushik Wasi$^{1}$,\\
Md Isfar Khan$^{2}$,
MD Manjurul Ahsan$^{3}$
}

\affil{
$^1$Shahjalal University of Science and Technology, Sylhet, Bangladesh\\
\vspace{-2.5mm}
$^2$Florida A\&M University, Tallahassee, FL, United States\\
\vspace{-2.5mm}
$^3$University of Oklahoma, Norman, OK, United States
}

\begin{document}

\input{sections/abstract}

\maketitle
\begin{abstract}
\end{abstract}

\vspace{3mm}
\input{sections/content}

\clearpage
\bibliography{main}

\appendix
\input{sections/appendix}
\appendix
\end{document}

%% file: macro.tex
\newcommand{\x}{\mathbf{x}}

\definecolor{blanchedalmond}{rgb}{1.0, 0.92, 0.8}
\definecolor{carmine}{rgb}{0.59, 0.0, 0.09}
\definecolor{lightblue}{rgb}{0.22,0.45,0.70}%

\renewcommand{\mathbf}{\boldsymbol}

\makeatletter
\def\Ddots{\mathinner{\mkern1mu\raise\p@
\vbox{\kern7\p@\hbox{.}}\mkern2mu
\raise4\p@\hbox{.}\mkern2mu\raise7\p@\hbox{.}\mkern1mu}}
\makeatother

\definecolor{amaranth}{rgb}{0.9, 0.17, 0.31}
\definecolor{antiquebrass}{rgb}{0.8, 0.58, 0.46}
\definecolor{antiquefuchsia}{rgb}{0.57, 0.36, 0.51}
\definecolor{chromeyellow}{rgb}{0.31, 0.47, 0.26}

%% file: sections/abstract.tex
\begin{abstract}
\textbf{Abstract:} Sleep quality is influenced by a complex interplay of behavioral, environmental, and psychosocial factors, yet most computational studies focus mainly on predictive risk identification rather than actionable intervention design. Although machine learning models can accurately predict subjective sleep outcomes, they rarely translate predictive insights into practical intervention strategies.
To address this gap, we propose a personalized predictive-prescriptive framework that integrates interpretable machine learning with mixed-integer optimization. A supervised classifier trained on survey data predicts sleep quality, while SHAP-based feature attribution quantifies the influence of modifiable factors. These importance measures are incorporated into a mixed-integer optimization model that identifies minimal and feasible behavioral adjustments, while modelling resistance to change through a penalty mechanism.
The framework achieves strong predictive performance, with a test F1-score of 0.9544 and an accuracy of 0.9366. Sensitivity and Pareto analyses reveal a clear trade-off between expected improvement and intervention intensity, with diminishing returns as additional changes are introduced. At the individual level, the model generates concise recommendations, often suggesting one or two high-impact behavioral adjustments and sometimes recommending no change when expected gains are minimal.
By integrating prediction, explanation, and constrained optimization, this framework demonstrates how data-driven insights can be translated into structured and personalized decision support for sleep improvement.

\vspace{0.5cm}

\coloremojicode{1F4C5} \textbf{Date}: March 15, 2026

\coloremojicode{1F4E7} \textbf{Correspondence}: Mahfuz Ahmed Anik~(\href{mailto:mahfuz34@student.sust.edu}{mahfuz34@student.sust.edu})

\coloremojicode{1F4BB} \textbf{Keywords}: Sleep Health Analytics, Mixed-Integer Optimization,\\ Prescriptive Analytics, Explainable AI, Personalized Decision Support


\end{abstract}

%% file: sections/content.tex
\section{Introduction}\label{sec1}

Sleep health is increasingly recognized as a central concern in public health and educational research, particularly among university and college students for whom chronic sleep insufficiency impairs learning capacity, emotional regulation, and long-term well-being \citep{milojevich2016sleep}. Student life concentrates multiple interacting pressures, including dense academic schedules, social obligations, part-time employment, shared housing, and excessive screen exposure, which collectively normalize irregular sleep routines and disrupt nighttime rest \citep{ming2011sleep}. Over time, these structural pressures transform seemingly individual lifestyle choices into systemic patterns of deprivation, positioning students among the most sleep-vulnerable populations globally \citep{hershner2014causes}. Large-scale surveys and localized investigations consistently report elevated rates of poor sleep within student populations, suggesting a persistent issue that awareness-based campaigns alone have not resolved \citep{dietrich2016effectiveness}. In certain contexts, the prevalence is even more severe; recent evidence from Bangladesh indicates that more than 90\% of university students report poor sleep quality \citep{ahmed2024association}. Such findings underscore that sleep disturbance among students is not an isolated behavioral anomaly but a structurally embedded condition within academic and social environments.

Conventional responses typically emphasize sleep hygiene education, promoting regular bedtimes, light exposure control, and environmental optimization \citep{migliaccio2024boost}. Although grounded in empirical sleep science, these recommendations often overlook the situational constraints students face. Many guidelines implicitly assume the feasibility of substantial lifestyle restructuring, yet financial, temporal, spatial, and social limitations frequently render such reforms impractical \citep{tomy2019barriers}. Consequently, students may understand the importance of healthy sleep while remaining unable to implement or sustain recommended changes. This disconnect reveals a central limitation of prevailing approaches: they prescribe what should change without systematically accounting for how much change is realistically attainable. Addressing sleep deprivation in young populations therefore requires not simply more information, but a reorientation toward intervention strategies that begin with behavioral constraints and design solutions within them.

Despite substantial advances in understanding sleep health and formulating evidence-based guidelines, a critical gap remains: most recommendations are delivered as generic checklists or broad prescriptions, offering limited guidance on selecting a small, feasible set of adjustments tailored to an individual student’s baseline sleep profile. This limitation is consequential because interventions that overlook the burden of behavioral change—including cognitive effort, routine disruption, and practical constraints—often experience low adherence, even when clinically appropriate \citep{rodriguez2019patient,burgess2017determinants}. Selecting intervention components is inherently a combinatorial and prescriptive task, requiring consideration of heterogeneous baselines, compatibility constraints among candidate actions, and realistic limits on the number of changes a student can implement. Prescriptive analytics has increasingly leveraged optimization-based techniques to translate predictive insights into actionable decisions across healthcare and related domains \citep{lepenioti2020prescriptive}. However, such approaches rarely accommodate the distinctive structure of sleep interventions, where the objective is incremental, minimally disruptive behavioral improvement rather than unconstrained outcome maximization \citep{mintz2023behavioral}.

Existing personalized sleep interventions frequently rely on one-size-fits-all assumptions or purely statistical predictions that neglect interaction effects among interventions \citep{saruhanjan2021psychological}, student-specific constraints \citep{liang2024co}, and explicit trade-offs between expected benefit and behavioral burden \citep{garbarino2024revolutionizing}. Consequently, they often produce recommendations that are difficult to operationalize, fail to prioritize feasible action sets, and insufficiently reflect individual circumstances. Moreover, many approaches overlook the complex interplay among physiological, environmental, and social determinants of sleep behavior, further limiting practical effectiveness \citep{philip2024sante,grandner2019social}. Collectively, these limitations underscore the inadequacy of current methodologies for generating modest, implementable, and context-sensitive sleep recommendations. Bridging this gap requires a framework capable of systematically balancing predictive accuracy, behavioral feasibility, and personalized constraint-aware intervention design.

\begin{figure*}[t] 
\centering
\includegraphics[width=\textwidth]{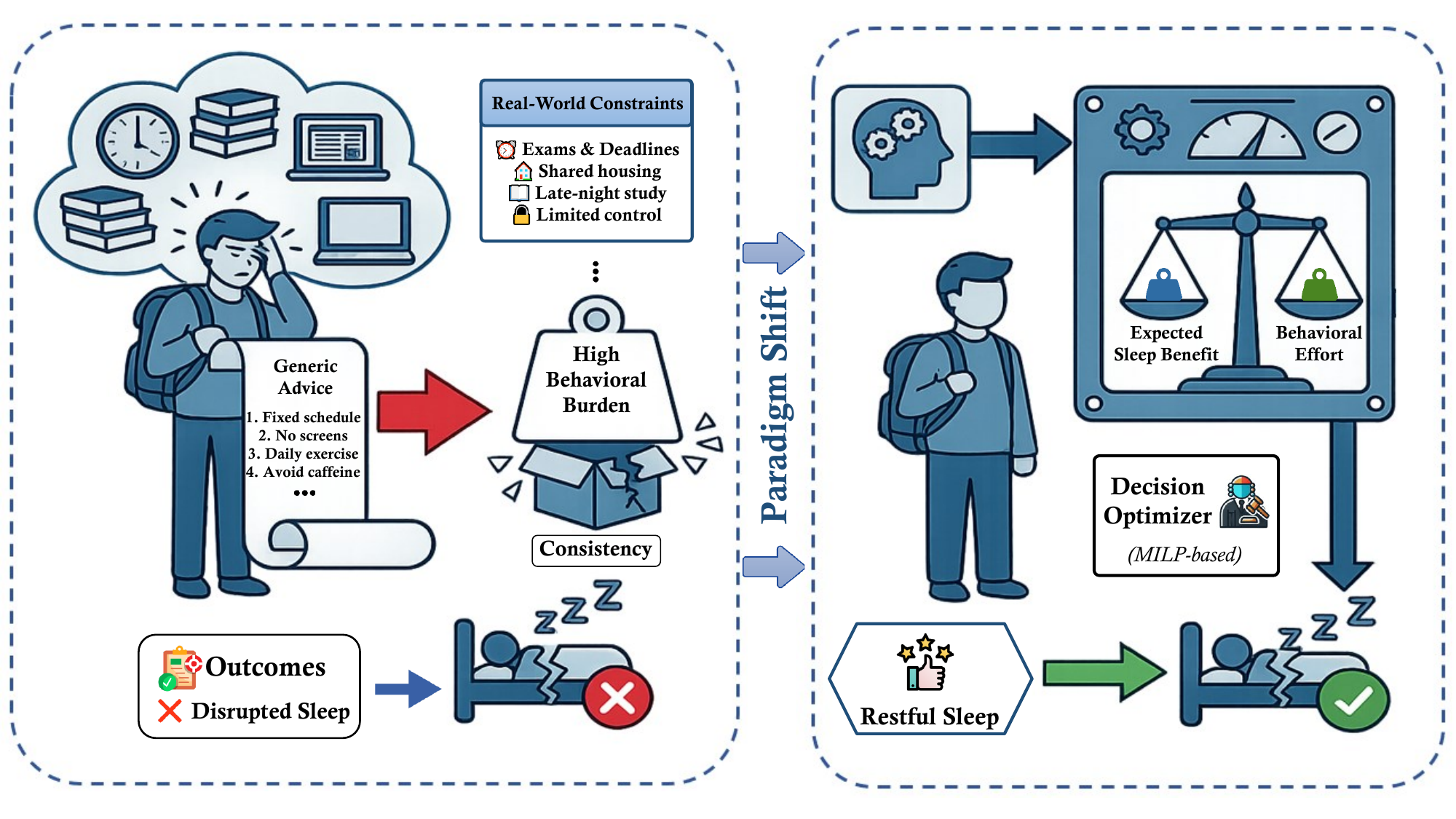}
\caption{Generic sleep advice imposes high behavioral burden under real-world constraints, frequently resulting in poor follow-through and disrupted sleep. A minimal-change optimization framework instead delivers personalized, low-effort actions that promote more sustainable sleep improvement.}
\label{fig:intro}
\end{figure*}

To address the fragmentation between predictive modelling and actionable intervention design in sleep health research, we propose a personalized predictive–prescriptive framework that integrates interpretable machine learning with constrained optimization. The framework is designed to translate feature-level importance estimates into structured, feasible behavioral recommendations at the individual level. A supervised learning model is first trained to estimate subjective sleep quality from survey-based demographic, behavioral, and environmental variables. Model explanations are then derived using SHAP to quantify the relative contribution of modifiable factors. These importance measures are not treated solely as diagnostic outputs; instead, they are embedded directly within a Mixed-Integer Linear Programming (MILP) formulation that selects a minimal set of feasible behavioral adjustments for each individual. The optimization model is constructed to balance expected improvement in predicted sleep quality against behavioral disruption. By introducing an explicit resistance parameter, the framework allows systematic control over intervention intensity and reflects heterogeneity in individual willingness to change. Rather than prescribing idealized sleep states or enforcing uniform behavioral targets, the formulation emphasizes proportionate, incremental adjustments that remain interpretable and practically attainable. This design positions explainability as a structural component of decision-making rather than as a post hoc interpretive layer.

Our contributions are summarized as follows:

\begin{enumerate}

\item We develop an integrated predictive–prescriptive framework that links sleep quality prediction, SHAP-based explanation, and mixed-integer optimization within a unified decision-support architecture.

\item We formalize personalized sleep intervention design as a constrained optimization problem, explicitly modelling feasibility, single-step behavioral adjustments, and resistance to change through a tunable penalty parameter.

\item We demonstrate how feature attribution measures derived from explainable machine learning can be operationalized as optimization coefficients, thereby establishing a methodological bridge between interpretability and prescriptive analytics in lifestyle intervention contexts.
\end{enumerate}


The remainder of this paper is organized as follows. Section~\ref{sec2} reviews relevant literature on sleep quality modelling, explainable artificial intelligence, and prescriptive analytics. Section~\ref{sec: data} describes the dataset, preprocessing procedures, and predictive modelling framework. Section~\ref{sec: ml_opt_framework} presents the proposed ML-informed optimization formulation and its integration with SHAP-based parameter estimation. Section~\ref{sec: exp_result} outlines the experimental design and reports empirical results. Section~\ref{sec: discussion} interprets the findings and examines practical implications and limitations. Finally, Section~\ref{sec: con} concludes the paper and highlights directions for future research.

\section{Literature Review}\label{sec2}

This section reviews prior research on sleep quality modelling, explainable artificial intelligence in health applications, and prescriptive analytics for decision support. We discuss established approaches to subjective sleep measurement and predictive modelling, examine advances in interpretability methods that enhance clinical transparency, and survey optimisation-driven frameworks that translate predictive insights into structured action. Finally, we identify methodological gaps that motivate the development of the proposed predictive prescriptive framework.

\subsection{Sleep Quality Measurement and Predictive Modelling}

Reliable modelling of sleep quality begins with understanding how it is measured. Subjective instruments, particularly the Pittsburgh Sleep Quality Index (PSQI), are widely adopted in clinical and research contexts, yet empirical evidence indicates that they capture experiential dimensions of sleep that do not map cleanly onto laboratory-based metrics such as polysomnography or actigraphy \citep{kaplan2017correlates}. Large-scale observational analyses show that canonical physiological markers explain only a limited proportion of variance in perceived sleep depth and restfulness, while contextual factors such as chronotype and workday structure can systematically influence PSQI summaries unless explicitly modelled \citep{kaplan2017correlates,pilz2018time}. These findings highlight the dual nature of PSQI as both clinically meaningful and behaviorally contextual, motivating predictive approaches that integrate demographic, environmental, and lifestyle variables.

Within this landscape, supervised learning pipelines have demonstrated that parsimonious feature sets can achieve strong classification performance when guided by principled selection strategies \citep{warunlawan2023identification}. Techniques such as minimum redundancy maximum relevance applied across multiple classifiers identify physiological and behavioural indicators—including blood pressure, body mass index, and physical activity—as consistently predictive features \citep{warunlawan2023identification}. Multimodal studies incorporating heart rate variability and skin temperature further report high accuracy under controlled experimental conditions \citep{di2024predicting}. More recent research emphasizes personalization and contextual modelling, where within-subject longitudinal analyses uncover individual-specific behavioural patterns linked to sleep outcomes \citep{upadhyay2020personalized}, and hybrid systems combining machine learning forecasts with large language models aim to transform predictions into explanation-oriented behavioural guidance \citep{corda2024context}.

Parallel intervention and clinical investigations demonstrate that modifiable lifestyle and environmental factors can yield measurable improvements in PSQI scores across diverse populations \citep{yu2024effect,li2025effects,bullock2020optimizing,calvo2021prevalence,botella2023evaluating}. Collectively, this literature indicates that while subjective sleep quality is only partially explained by physiological markers, predictive models leveraging behavioural and contextual features can achieve meaningful performance. The evidence further suggests that incorporating personalization and interpretability is essential for translating predictive insights into actionable and behaviorally realistic sleep interventions.

\subsection{Explainable AI for Sleep and Health Applications}

Recent advances in sleep quality prediction have shifted emphasis from predictive accuracy toward interpretability and clinical usability. Systems-level analyses of digital sleep health highlight that robust preprocessing of heterogeneous signals, principled sensor fusion, and transparent model outputs are essential for translating algorithmic predictions into clinically meaningful tools \citep{perez2020future}. The proliferation of wearable sensing technologies has enabled multimodal monitoring beyond laboratory settings, where integrated architectures combining photoplethysmography, differential air pressure, and tri-axial acceleration achieve performance comparable to expert annotation while offering confidence estimates and saliency-based explanations for clinical review \citep{rossi2023sleep}. Similarly, compact EEG and PPG-based pipelines demonstrate how gradient-boosted models augmented with SHAP-guided clustering and symbolic rule extraction can reveal physiologically interpretable stage-specific signatures under reduced hardware configurations \citep{xu2025interpretable}.

Parallel methodological developments advocate hybrid modelling strategies that retain deep learning’s representational power while incorporating post hoc attribution mechanisms and symbolic summaries to enhance transparency. Tree-ensemble models trained on wearable signals have shown meaningful discrimination of perceived sleep quality while producing medically coherent feature attributions that facilitate clinical adoption \citep{moebus2024personalized}. Reviews and empirical analyses further demonstrate that SHAP and related explanation techniques can uncover physiologically grounded contributors to sleep and stress states and argue for their systematic inclusion in validation and reporting protocols \citep{barati2024interpretable,chintalapati2024enhancing}. At larger scales, modular architectures integrating demographic sequence modelling, flexible channel analysis, and attention mechanisms illustrate that generalizability and interpretability can coexist, yielding clinician-oriented visualizations validated across multi-ethnic cohorts \citep{hu2025transparent}. Nevertheless, methodological assessments caution that cross-cohort robustness and resilience to device heterogeneity remain central challenges for real-world deployment \citep{nasir2023transformative,perez2020future}.

\subsection{From Prediction to Prescription: ML-Informed optimization and Decision Support}

As interpretability becomes a structural component of sleep analytics, attention increasingly turns toward systems that not only explain predictions but also recommend feasible actions. Hybrid decision-support architectures illustrate this shift. For example, machine learning classifiers augmented with fuzzy expert layers have been used to generate risk indicators and diagnostic suggestions for obstructive sleep apnea, embedding clinical heuristics within transparent rule structures while maintaining useful discriminative performance \citep{casal2023design}. Such designs demonstrate how symbolic reasoning can serve as a safeguard when purely data-driven models face distributional shifts. Beyond individual diagnosis, environmental and organisational determinants of sleep highlight the relevance of prescriptive levers at broader scales. Reviews of indoor environmental quality identify consistent associations between thermal comfort, nocturnal noise, evening light exposure and sleep continuity, positioning building-level control as an actionable intervention mechanism \citep{yasmeen2025exploring}. Organisational fatigue research similarly shows that work scheduling and task allocation may exert stronger influence on sleep deficiency than residential factors, suggesting that operational design itself can function as a prescriptive tool \citep{jameson2023sleep}. At the same time, hospital-based studies reveal that perceived improvements in sleep do not always correspond to measurable physiological gains, underscoring the complexity of evaluating intervention impact in real settings \citep{thomas2012sleep}. Formal optimization frameworks offer a structured pathway for translating predictive insights into operational decisions. Surveys in healthcare operations research document how scheduling, forecasting and resource allocation models deliver measurable improvements across service systems \citep{rais2011operations}. In sleep and alertness management, biologically informed optimization of sleep–work schedules reduces predicted impairment and accelerates recovery from chronic restriction, although existing calibrations remain focused primarily on young, healthy cohorts \citep{vital2021optimal}. Parallel developments in mental health decision support show that explainable classifiers can generate actionable recommendations, while also exposing the risks of overfitting and dataset-specific artefacts \citep{payne2025enhancing}. Neuro-fuzzy systems provide an alternative interpretable paradigm by encoding decision logic in human-readable rules and explicitly modelling uncertainty, though empirical demonstrations often rely on limited pilot samples \citep{cheriyan2024adaptive}. Together, these strands of research indicate that prescriptive sleep systems require more than predictive accuracy; they demand structured optimization mechanisms that remain interpretable, behaviourally realistic, and robust under practical constraints.

\subsection{Research Gap Analysis}

While machine learning approaches have demonstrated promising performance in sleep quality classification \citep{warunlawan2023identification,di2024predicting}, the majority of existing studies remain confined to predictive risk estimation. These works primarily determine whether sleep quality is poor or satisfactory, but do not formalize the subsequent decision process required to identify which specific behavioural or environmental adjustments should be implemented. Even in personalized modelling settings \citep{upadhyay2020personalized,corda2024context}, recommendations are typically derived from correlational patterns or feature rankings rather than from structured optimization frameworks that explicitly account for feasibility and behavioural constraints. Parallel advances in explainable artificial intelligence have strengthened interpretability in sleep and health analytics \citep{moebus2024personalized,barati2024interpretable,chintalapati2024enhancing}. However, attribution methods such as SHAP are predominantly used for post hoc explanation rather than as integral components of decision mechanisms. Feature importance measures are seldom operationalized as formal decision coefficients within constrained optimization models, and interactions among actionable factors are rarely considered within a unified prescriptive architecture. As a consequence, prediction, explanation, and intervention design are often treated as sequential stages rather than as a structurally integrated framework. Prescriptive and optimization-based approaches in related health and operational domains illustrate the feasibility of translating predictive insights into structured action \citep{rais2011operations,vital2021optimal,casal2023design}. Yet within sleep health research, the methodological integration of machine learning, explainability, and individual-level optimization remains limited. 

To address these limitations, we proposed a structurally integrated framework that connects prediction, explanation, and prescriptive decision-making within a unified architecture. Rather than treating risk estimation and interpretability as terminal analytical outputs, predictive importance measures are positioned as inputs to a formal decision process that explicitly accounts for feasibility and behavioural constraints. In doing so, the work moves beyond descriptive modelling toward a decision-oriented perspective, in which sleep improvement is framed as an optimization problem grounded in empirical evidence. This perspective responds directly to the methodological fragmentation observed in prior research and establishes a coherent pathway from predictive insight to actionable guidance.

\section{Data and Predictive Modeling } \label{sec: data}
Effective prescriptive decision-making depends on accurate and interpretable prediction. To ensure that subsequent optimization is grounded in empirical evidence, the data generation process and predictive modeling framework are first established. The next section details the survey dataset, preprocessing methodology, and model benchmarking procedures that support the proposed intervention architecture.

\subsection{Dataset Description and Preprocessing}
\subsubsection{Survey design}

\noindent \textbf{Data Collection Framework.}
The dataset was collected through a structured, self-administered online survey distributed via Google Forms. The target population consisted of young individuals aged below 30 years. The instrument was designed to capture multidimensional determinants of sleep quality while maintaining compatibility with predictive modeling and prescriptive optimization. Emphasis was placed on balancing comprehensiveness with interpretability.

\noindent \textbf{Demographic and Residential Context.}
The first section collected demographic and residential attributes, including age, gender, living arrangement, housing type, and sleeping environment characteristics. These variables provide contextual background and may indirectly influence sleep through environmental or lifestyle conditions.

\noindent \textbf{Behavioral and Lifestyle Factors.}
The second section focused on behavioral variables associated with sleep quality, including screen exposure before bedtime, caffeine intake, physical activity level, academic or work-related workload, and pre-sleep routines. Most variables were measured using ordinal scales to reflect realistic behavioral variation.

\noindent \textbf{Environmental and Psychological Assessment.}
Participants evaluated bed comfort, lighting, room quietness, and ventilation using five-point Likert scales \citep{allen2007likert}. Psychological and stress-related variables—such as perceived academic pressure, financial stress, and recurring physical discomfort—were incorporated to capture non-physical determinants of sleep.

\noindent \textbf{Sleep Quality Measurement.}
Sleep quality was measured using the Pittsburgh Sleep Quality Index (PSQI) \citep{smyth1999pittsburgh}, a validated instrument assessing subjective sleep quality over a one-month period. In total, 418 completed responses formed the raw dataset for subsequent preprocessing and modeling.

\subsubsection{Data Preprocessing and Feature Engineering}
Following initial data cleaning, a structured preprocessing pipeline was applied to transform the raw survey responses into a machine-learning-ready format while preserving behavioral interpretability. 

\noindent \textbf{Encoding Strategy.}
A structured preprocessing pipeline was applied to transform raw survey responses into a machine-learning-ready format while preserving behavioral interpretability. Binary variables were mapped to indicator values, while ordinal variables—such as screen time, physical activity, workload, sleep schedule consistency, and environmental ratings—were encoded using predefined ordinal mappings reflecting meaningful behavioral gradations. Nominal variables without inherent ordering were encoded using label encoding to ensure compatibility with tree-based models.

\noindent \textbf{Outlier Treatment.}
Outliers in numerical features were detected using the interquartile range (IQR) criterion and treated via capping to reduce the influence of extreme values while preserving sample size.

\noindent \textbf{Feature Engineering.}
Several engineered features were introduced to enhance representational capacity. Continuous variables such as age and body mass index (BMI) were discretized into clinically meaningful categories. Composite indices were constructed to summarize correlated constructs, including a sleep environment score (aggregating bed comfort, lighting, quietness, and ventilation) and a lifestyle score (combining screen exposure and physical activity). A binary stress indicator and a poor sleep habits score were also derived to capture higher-level behavioral patterns.

\noindent \textbf{Feature Selection.}
To identify the most informative predictors, multiple feature selection techniques were applied, including univariate F-tests, mutual information analysis, recursive feature elimination using random forests, and tree-based importance rankings. Features consistently selected across methods were retained, resulting in a final set of 15 predictors.

\begin{figure*}[htbp] 
\centering
\includegraphics[width=\textwidth]{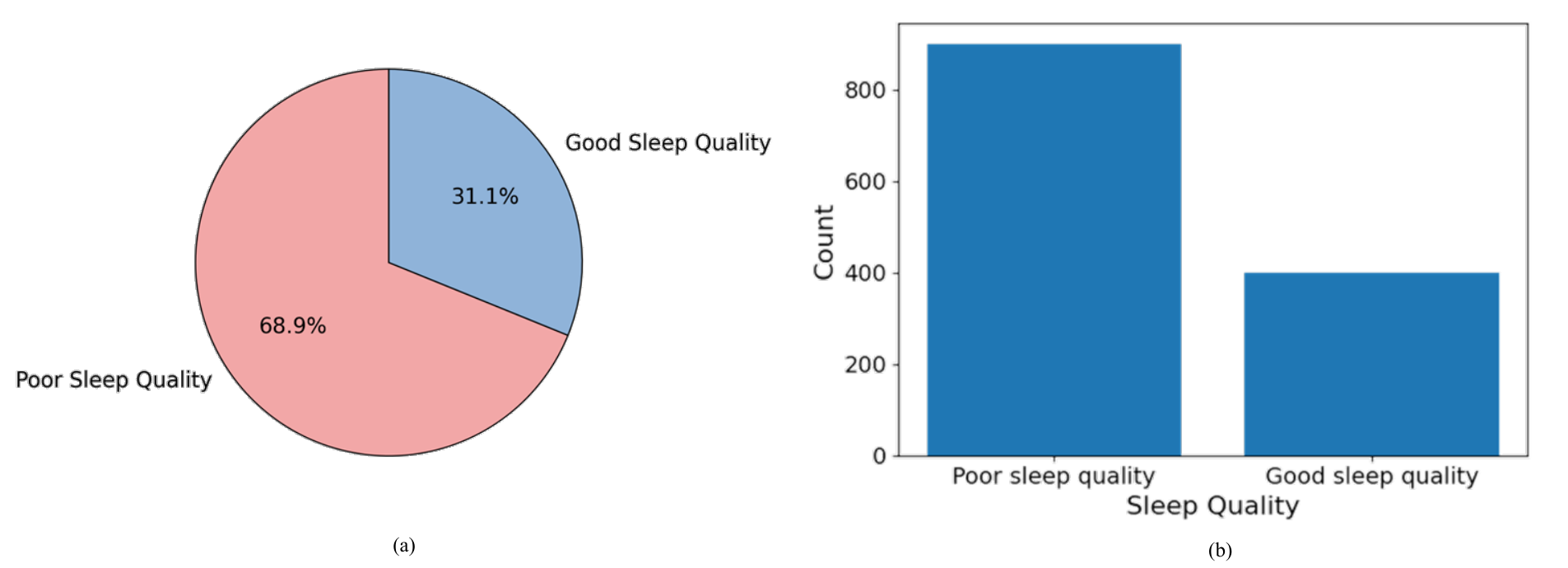}
\caption{(a) Class distribution of sleep quality; (b) Sample counts for each sleep quality class after data augmentation.}
\label{fig:eda_1}
\end{figure*}

\begin{figure*}[htbp] 
\centering
\includegraphics[width=\textwidth]{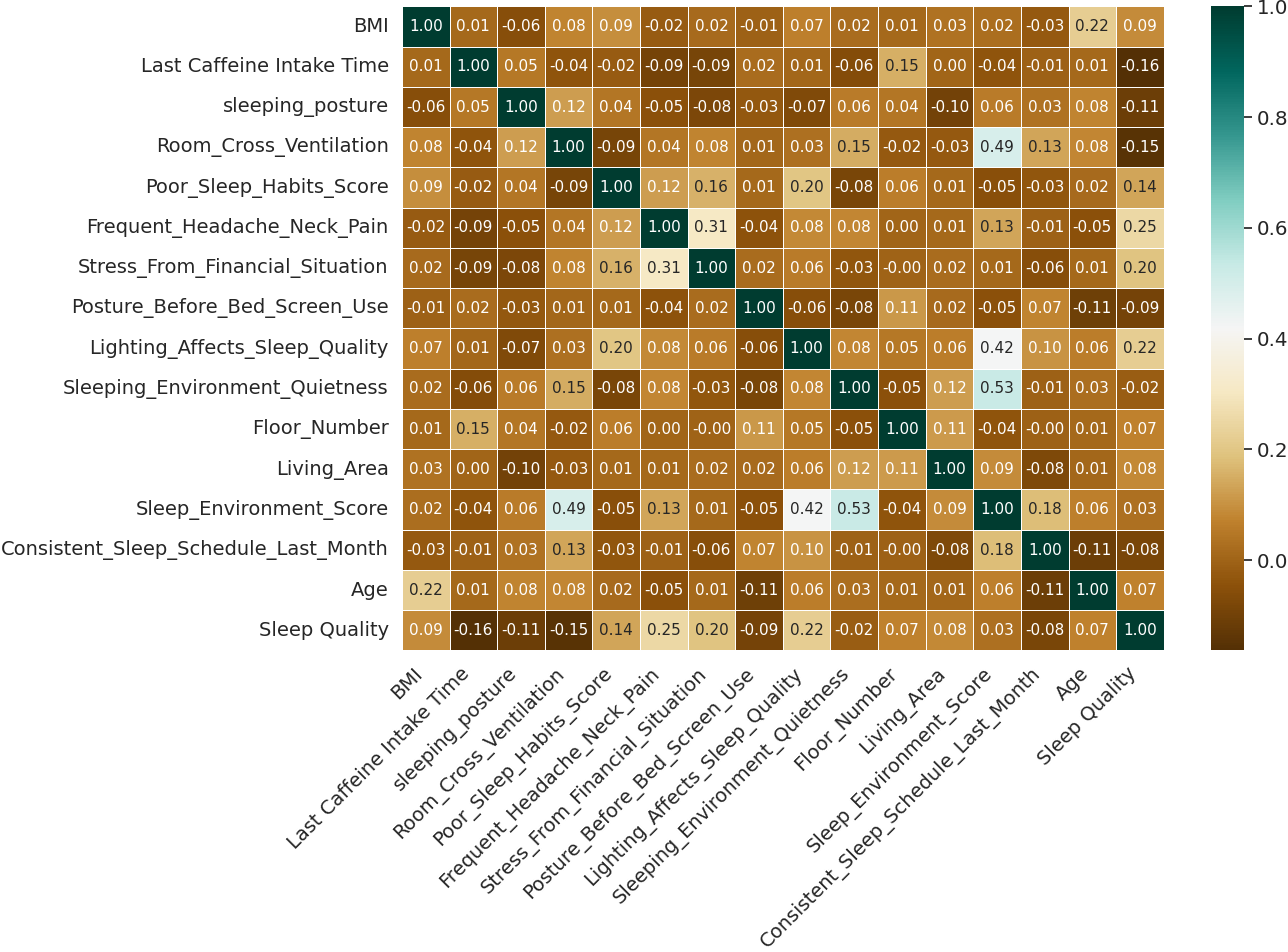}
\caption{Pearson correlation matrix of sleep quality and associated features}
\label{fig:eda_2}
\end{figure*}

\paragraph{Exploratory Data Analysis.}
Exploratory data analysis provides insight into the structural properties of the dataset prior to modeling. As illustrated in Fig.~\ref{fig:eda_1}, the original survey responses exhibit a noticeable class imbalance, with poor sleep quality more prevalent than good sleep quality. After data augmentation, the class distribution becomes more balanced, although moderate imbalance remains. In total, the original dataset consisted of 418 responses, which were expanded to 1,339 samples with 15 features to improve model stability and representation across classes. The final distribution comprises 922 instances labeled as good sleep quality and 417 as poor sleep quality, corresponding to an imbalance ratio of 0.45. Figure~\ref{fig:eda_2} presents the Pearson correlation matrix between sleep quality and associated behavioral and environmental variables. The heatmap highlights meaningful associations, particularly for sleep environment conditions, screen use before bedtime, and stress-related factors. These observed relationships provide empirical support for incorporating composite behavioral features and interaction-aware representations in the subsequent predictive modeling and optimization stages. The processed dataset formed the basis for all downstream analyses.

\subsubsection{Actionable and Non-Actionable Variables}
To ensure that the proposed optimization framework yields feasible and practically meaningful recommendations, the study explicitly distinguishes between actionable and non-actionable variables. Non-actionable variables include demographic, physiological, and contextual attributes such as age, gender, body mass index, academic workload, and residential characteristics, which may strongly influence sleep quality but cannot be directly modified through short-term interventions. These variables are therefore retained exclusively within the predictive modeling stage to enhance estimation accuracy. In contrast, actionable variables correspond to modifiable behavioral and environmental factors that individuals can realistically adjust, including screen use before bedtime, timing of caffeine intake, sleep schedule regularity, sleeping posture, and bedroom conditions related to lighting, quietness, and ventilation. All actionable variables were encoded using ordinal scales that reflect incremental and behaviorally realistic changes, enabling discrete intervention modeling. Table~\ref{tab:actionable_variables} summarizes the classification of variables by actionability and clarifies their respective roles within the framework. This explicit separation underpins the subsequent mixed-integer linear programming formulation by restricting decision variables to actionable factors only, thereby preserving interpretability, feasibility, and alignment with real-world decision-making constraints.

\begin{table}[htbp]
\centering
\caption{Classification of variables by actionability and role in the framework}
\label{tab:actionable_variables}
\begin{tabular}{p{2.5cm} p{5.8cm} p{3.5cm}}
\hline
\textbf{Variable Category} & \textbf{Examples} & \textbf{Role in Framework} \\
\hline
Non-actionable &
Age, Gender, BMI, Academic workload, Housing context &
Used only for sleep quality prediction \\

Actionable &
Screen use before sleep, Caffeine intake timing, Sleep schedule consistency, \\
& Sleeping posture, Lighting, Quietness, Ventilation &
Decision variables in optimization model \\
\hline
\end{tabular}
\end{table}

\subsection{Sleep Quality Prediction and Explainability}

\subsubsection{Benchmarking of Machine Learning Models}

To assess the effectiveness of different predictive approaches for sleep quality classification, a set of widely used machine learning models was systematically benchmarked. The evaluated models include XGBoost \citep{chen2016xgboost}, LightGBM \citep{ke2017lightgbm}, Gradient Boosting \citep{friedman2001greedy}, Extra Trees \citep{geurts2006extremely}, Random Forest \citep{breiman2001random}, and a multilayer perceptron (MLP) \citep{tolstikhin2021mlp}, representing both ensemble-based and neural network-based learning paradigms. Model performance was evaluated using multiple standard metrics, including accuracy, precision, recall, and F1-score, with particular emphasis placed on the F1-score due to the moderate class imbalance present in the dataset. Table~\ref{tab:ml_benchmark} summarizes the predictive performance of all models on the test set, along with corresponding training and validation F1-scores. Overall, tree-based ensemble methods consistently outperformed the neural network baseline, highlighting their suitability for structured, tabular survey data. Among all evaluated models, XGBoost achieved the highest test F1-score (0.9544), while also maintaining strong precision and recall values. Importantly, the gap between training, validation, and test performance for XGBoost remained small, indicating stable generalization and limited overfitting. Although other ensemble models such as LightGBM and Gradient Boosting demonstrated comparable performance, XGBoost provided the best overall balance between predictive accuracy, robustness, and consistency across data splits. Given its superior performance and compatibility with model-agnostic explainability techniques, XGBoost was selected as the primary predictive model for subsequent explainability analysis using SHAP and for informing the optimization-based intervention framework. This selection ensures that the prescriptive decisions derived in later stages are grounded in a high-performing and well-generalized predictive model.

\begin{table}[t]
\centering
\caption{Performance metrics of evaluated machine learning models}
\label{tab:ml_benchmark}
\begin{tabular}{lcccccc}
\hline
\textbf{Model} & \textbf{Test F1} & \textbf{Accuracy} & \textbf{Precision} & \textbf{Recall} & \textbf{Val F1} & \textbf{Train F1} \\
\hline
XGBoost & \textbf{0.9544} & \textbf{0.9366} & \textbf{0.9468} & 0.9622 & 0.9579 & 0.9915 \\
LightGBM & 0.9526 & 0.9328 & 0.9282 & 0.9784 & \textbf{0.9630} & 0.9938 \\
Gradient Boosting & 0.9496 & 0.9291 & 0.9323 & 0.9676 & 0.9479 & \textbf{0.9954} \\
Extra Trees & 0.9476 & 0.9254 & 0.9188 & 0.9784 & 0.9524 & 0.9735 \\
Random Forest & 0.9455 & 0.9216 & 0.9100 & \textbf{0.9838} & 0.9430 & 0.9817 \\
Neural Network (MLP) & 0.8241 & 0.7388 & 0.7700 & 0.8865 & 0.8700 & 0.8458 \\
\hline
\end{tabular}
\end{table}

\subsubsection{XGBoost Configuration and Tuning}

XGBoost is a gradient boosting framework that constructs an additive ensemble of decision trees in a stage-wise manner \citep{chen2016xgboost}. Given a training dataset $\{(x_i, y_i)\}_{i=1}^{n}$, the model prediction at iteration $t$ is expressed as

\begin{equation}
\hat{y}_i^{(t)} = \sum_{k=1}^{t} f_k(x_i),
\end{equation}

where each $f_k$ represents a regression tree belonging to the functional space $\mathcal{F}$ of decision trees. The model is trained by minimizing a regularized objective function defined as

\begin{equation}
\mathcal{L}^{(t)} = \sum_{i=1}^{n} \ell \big( y_i, \hat{y}_i^{(t-1)} + f_t(x_i) \big) + \Omega(f_t),
\end{equation}

where $\ell(\cdot)$ denotes the differentiable loss function and $\Omega(f_t)$ is a regularization term that penalizes model complexity. The regularization component is given by

\begin{equation}
\Omega(f) = \gamma T + \frac{1}{2}\lambda \| w \|^2 + \alpha \| w \|_1,
\end{equation}

where $T$ is the number of leaves in the tree, $w$ denotes leaf weights, and $\lambda$ and $\alpha$ control $L_2$ and $L_1$ regularization, respectively. This formulation enhances generalization performance by controlling tree complexity and preventing overfitting.

\begin{table}[t]
\centering
\caption{XGBoost hyperparameter configuration and search space}
\label{tab:xgb_hyperparams}
\begin{tabular}{lcc}
\hline
\textbf{Hyperparameter} & \textbf{Search Range} & \textbf{Selected Value} \\
\hline
Number of trees ($n\_estimators$) & \{50, 100\} & 100 \\
Learning rate ($\eta$) & \{0.01, 0.05, 0.1\} & 0.05 \\
Maximum depth ($\text{max\_depth}$) & \{2, 3, 4\} & 3 \\
Subsample ratio ($\text{subsample}$) & \{0.6, 0.8\} & 0.8 \\
Column sample ratio ($\text{colsample\_bytree}$) & \{0.6, 0.8\} & 0.8 \\
Minimum child weight ($\text{min\_child\_weight}$) & \{3, 5\} & 3 \\
L1 regularization ($\alpha$) & \{0.1, 1.0\} & 0.1 \\
L2 regularization ($\lambda$) & \{0.1, 1.0\} & 1.0 \\
\hline
\end{tabular}
\end{table}

Hyperparameters were tuned using grid search over predefined ranges. The explored search space and the final selected configuration are summarized in Table~\ref{tab:xgb_hyperparams}. Model selection was based on validation F1-score, prioritizing generalization stability and robustness to class imbalance. The final selected configuration was determined based on validation F1-score performance. Emphasis was placed on controlling model complexity while maintaining high predictive accuracy, ensuring that the resulting classifier remained stable and well-suited for downstream explainability and optimization stages.

\subsubsection{Model Explainability using SHAP}

To interpret the predictions of the selected XGBoost model and to identify the most influential factors driving sleep quality outcomes, SHapley Additive exPlanations (SHAP) were employed \citep{lundberg2017unified}. SHAP provides a unified, model-agnostic framework grounded in cooperative game theory, decomposing a model’s prediction into additive contributions from individual features. This enables both global and local interpretability while accounting for nonlinear effects and feature interactions, which are particularly relevant for tree-based ensemble models.

\begin{figure}[htbp]
\centering
\includegraphics[width=\linewidth]{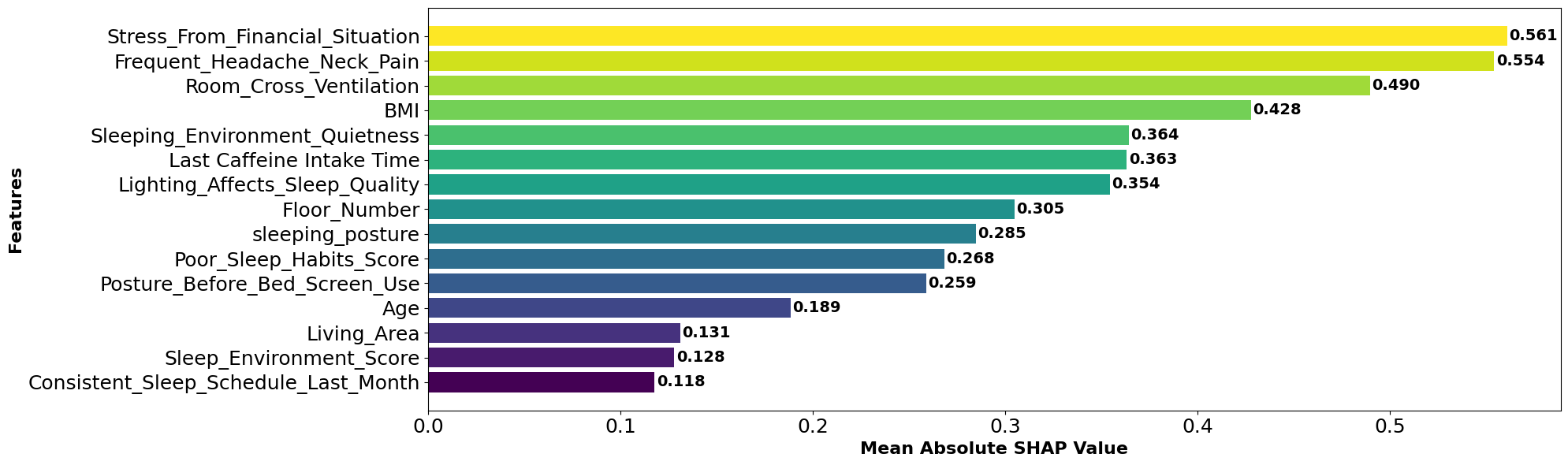}
\caption{Mean absolute SHAP values for the XGBoost sleep quality prediction model. Higher values indicate features with greater influence on model predictions.}
\label{fig:shap_importance}
\end{figure}

Figure~\ref{fig:shap_importance} presents the mean absolute SHAP values for the most influential features, summarizing their average contribution to the model’s output across the evaluation dataset. Higher SHAP values indicate features with greater impact on predicted sleep quality. The results reveal that stress-related variables, such as financial stress and frequent headaches or neck pain, exert the strongest influence on sleep quality predictions. Environmental factors, including room cross-ventilation, nighttime quietness, and lighting conditions, also emerge as highly influential, underscoring the importance of the sleep environment. Behavioral factors, such as screen use before bedtime, caffeine intake timing, sleeping posture, and sleep schedule consistency, exhibit moderate but meaningful contributions. Beyond interpretability, SHAP plays a critical role in bridging prediction and prescription within the proposed framework. Rather than using SHAP solely as a diagnostic tool, the mean absolute SHAP values associated with actionable variables are later leveraged as importance weights in the optimization objective, as detailed in Section~\ref{sec:shap-based estimation}. This ensures that intervention recommendations prioritize changes that are empirically supported by the predictive model, while remaining transparent and interpretable. By grounding the optimization parameters in SHAP-based explanations, the framework maintains consistency between predictive insights and prescriptive decision-making, enabling principled and data-driven personalized sleep interventions.

\section{ML-Informed Optimization Framework}\label{sec: ml_opt_framework}

\begin{figure}[htbp]
\centering
\includegraphics[width=\linewidth]{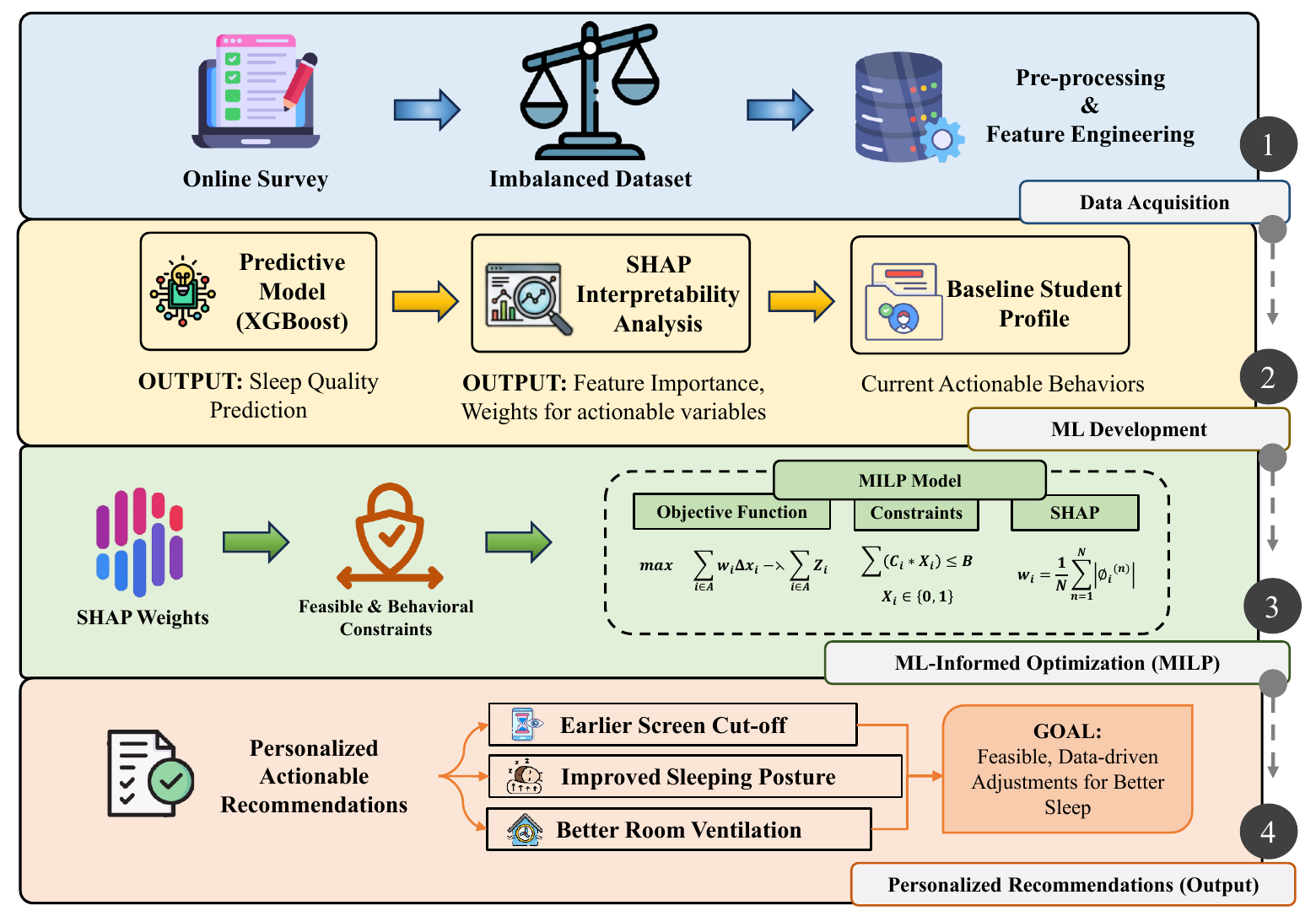}
\caption{Overview of the proposed predictive–prescriptive framework. Survey data are preprocessed and used to train an XGBoost model for sleep quality prediction. SHAP-based analysis derives importance weights for actionable variables, which are incorporated into a constrained mixed-integer linear programming (MILP) model. The optimization produces personalized, feasible, and minimally disruptive sleep improvement recommendations.}
\label{fig:framework}
\end{figure}

Building upon the predictive and explainability components established earlier, the framework now transitions from estimation to decision-making. As illustrated in Figure~\ref{fig:framework}, SHAP-derived feature importance measures are integrated with a constrained optimization model to translate predictive insights into personalized interventions. This section formalizes the resulting ML-informed mixed-integer linear programming formulation and defines the decision structure underlying individualized sleep improvement recommendations.

\subsection{Problem Definition}\label{subsec: problem}
We consider a personalized intervention optimization problem in which the objective is to support individual students in improving sleep quality through a small number of feasible, behaviorally realistic adjustments. Each student is characterized by a fixed baseline profile derived from survey data, capturing current sleep-related behaviors and environmental conditions. Rather than seeking to prescribe generic advice or enforce target outcomes, the problem is framed as a decision-support task that determines which controllable factors, if adjusted, are expected to yield meaningful improvements for a specific individual. Central to this formulation is a minimal-change philosophy. The optimization does not attempt to overhaul a student’s lifestyle or force convergence to an idealized sleep state. Instead, it identifies the smallest set of actionable modifications whose anticipated benefit justifies the effort required to implement them. This perspective reflects practical behavioral constraints, acknowledges heterogeneity in individual readiness to change, and aligns with the role of optimization as a prescriptive tool that balances improvement against disruption. Consequently, the problem is defined not as maximizing sleep quality at all costs, but as selecting interventions that are both effective and proportionate, given the student’s existing profile.

\subsection{ML-Informed Mixed-Integer Linear Programming Formulation}

To operationalize the personalized intervention problem defined in Section \ref{subsec: problem}, we formulate a mixed-integer linear programming (MILP) model that identifies a set of feasible behavioral and environmental adjustments tailored to an individual student \citep{artigues2014mixed}. The formulation explicitly balances the expected benefit of each intervention against the burden associated with behavioral change, while preserving interpretability and computational tractability.

\subsubsection{Decision Variables}

Let \( \mathcal{A} \) denote the set of actionable variables retained for optimization. For each \( i \in \mathcal{A} \), we define two decision variables.

First, an integer-valued change variable
\begin{equation}
\Delta x_i \in \mathbb{Z}_{\ge 0}, 
\quad \forall i \in \mathcal{A}
\label{eq:delta_domain}
\end{equation}
represents the magnitude of improvement applied to variable \( i \). A value of \( \Delta x_i = 1 \) corresponds to a one-level improvement along the predefined ordinal scale of the variable, while \( \Delta x_i = 0 \) indicates no change from the baseline state.

Second, a binary activation variable
\begin{equation}
z_i \in \{0,1\}, 
\quad \forall i \in \mathcal{A}
\label{eq:z_domain}
\end{equation}

indicates whether an intervention is applied to variable \( i \). This auxiliary variable allows the model to penalize the number of distinct interventions independently of their magnitude, thereby supporting a minimal-change philosophy.

For each student, baseline values \( x_i^0 \) are treated as fixed parameters and are not decision variables.

\subsubsection{Objective Function}

The objective is to maximize the expected improvement in sleep quality induced by the selected interventions, while discouraging unnecessary or excessive behavioral changes. Let \( w_i \ge 0 \) denote the importance weight associated with actionable variable \( i \), derived from the mean absolute SHAP value of the trained predictive model (as detailed in Section 4.3). The objective function is defined as

\begin{equation}
\max \quad 
\sum_{i \in \mathcal{A}} w_i \Delta x_i 
- \lambda \sum_{i \in \mathcal{A}} z_i
\label{eq:objective}
\end{equation}

The first term approximates the marginal contribution of incremental improvements in actionable variables to sleep quality, leveraging the interpretability of SHAP values as additive importance measures. The second term introduces a penalty proportional to the number of activated interventions, where \( \lambda \ge 0 \) is a behavioral resistance parameter controlling the trade-off between predicted benefit and lifestyle disruption.

This formulation avoids enforcing explicit outcome targets and instead prioritizes proportionate, high-impact recommendations.

\subsubsection{Feasibility and Behavioral Constraints}

\paragraph{Bounds on Adjusted Variables}

Each actionable variable is restricted to remain within realistic and data-consistent bounds after intervention. Let \( \underline{x}_i \) and \( \overline{x}_i \) denote the minimum and maximum feasible values of variable \( i \), respectively. Then,

\begin{equation}
\underline{x}_i 
\le x_i^0 + \Delta x_i 
\le \overline{x}_i,
\quad \forall i \in \mathcal{A}
\label{eq:bounds}
\end{equation}

These bounds ensure that the optimization does not recommend infeasible or implausible states.

\paragraph{Unit-Change Restriction for Ordinal Variables}

To reflect behavioral realism and prevent abrupt lifestyle shifts, changes to ordinal variables are restricted to at most one level:

\begin{equation}
0 \le \Delta x_i \le 1,
\quad \forall i \in \mathcal{A}
\label{eq:unit}
\end{equation}

This constraint enforces gradual adjustment and supports interpretability of recommendations.

\paragraph{Linking Intervention Activation and Magnitude}

To ensure that the intervention penalty is incurred only when a variable is actively modified, the following linking constraint is imposed:

\begin{equation}
\Delta x_i \le z_i,
\quad \forall i \in \mathcal{A}
\label{eq:link}
\end{equation}

As a result, \( z_i = 1 \) if and only if an improvement is applied to variable \( i \).

\subsubsection{Variable Domains}

The decision variables are defined over the following domains:

\begin{equation}
\Delta x_i \in \mathbb{Z}_{\ge 0}, 
\quad
z_i \in \{0,1\},
\quad \forall i \in \mathcal{A}
\label{eq:domains}
\end{equation}

Together with the linear objective and constraints, this defines a mixed-integer linear program that can be solved efficiently using standard optimization solvers.

\subsubsection{Interpretation of the Optimal Solution}

The optimal solution yields a personalized intervention plan specifying which actionable variables should be adjusted and by how much. Each nonzero \( \Delta x_i \) corresponds to a concrete, one-level improvement in a specific behavior or environmental condition, while the structure of the objective function ensures that only interventions with sufficient expected benefit are selected. Importantly, the formulation allows for the possibility that no intervention is recommended when the anticipated gains do not justify behavioral disruption, reinforcing the decision-support nature of the model.

\subsection{SHAP-Based Parameter Estimation}
\label{sec:shap-based estimation}
The optimization model introduced in Section 4.2 relies on variable-specific parameters that quantify the expected contribution of each actionable intervention to sleep quality. These parameters are estimated using SHAP, which provides a principled, model-agnostic decomposition of a predictive model’s output into additive feature contributions. Leveraging SHAP enables a direct and interpretable link between the predictive and prescriptive components of the framework.

Let \( f(\cdot) \) denote the trained sleep quality prediction model and let \( \phi_i^{(n)} \) represent the SHAP value of actionable variable \( i \) for student \( n \). The SHAP value measures the marginal contribution of feature \( i \) to the model’s prediction relative to a baseline expectation, accounting for interactions with other features. While individual SHAP values are student-specific, the optimization model requires a stable, population-level estimate of relative importance to parameterize the objective function.

Accordingly, for each actionable variable \( i \in \mathcal{A} \), we define its optimization weight as the mean absolute SHAP value across the evaluation dataset:
\begin{equation}
w_i = \frac{1}{N} 
\sum_{n=1}^{N} 
\left| \phi_i^{(n)} \right|,
\quad \forall i \in \mathcal{A}
\label{eq:shap_weight}
\end{equation}

where \( N \) denotes the number of samples used for SHAP analysis. The use of absolute values captures the magnitude of influence irrespective of direction, reflecting the potential impact of adjusting a variable rather than the sign of its association in a specific instance. These weights serve as coefficients in the objective function of the MILP formulation and approximate the expected marginal improvement in sleep quality associated with a one-unit enhancement in the corresponding actionable variable. Importantly, the weights are derived from the same predictive model used to assess sleep quality, thereby ensuring internal consistency between prediction and prescription. The choice of mean absolute SHAP values is motivated by three considerations. First, SHAP values satisfy desirable axiomatic properties, including additivity and consistency, which support their use as surrogate marginal effects in optimization. Second, averaging across individuals yields robust population-level estimates that are less sensitive to idiosyncratic profiles, thereby stabilizing the behavior of the optimization model. Third, the resulting weights preserve interpretability, as larger values directly indicate variables whose incremental adjustments are expected to exert greater influence on predicted sleep quality.

\begin{table}[htbp]
\centering
\caption{SHAP-Derived Weights for Actionable Variables}
\label{tab:shap_weights}
\begin{tabular}{p{3.2cm} p{3.8cm} p{2.5cm} p{1.8cm}}
\hline
\textbf{Actionable Variable} & \textbf{Description} & \textbf{Mean Absolute SHAP Value (\(w_i\))} & \textbf{Normalized Weight (\(\tilde{w}_i\))} \\
\hline
Room cross-ventilation & Quality of air circulation in the sleeping environment & 0.490 & 0.160 \\
Nighttime quietness & Level of ambient noise during sleep & 0.364 & 0.119 \\
Lighting condition & Influence of lighting on sleep quality & 0.354 & 0.116 \\
Last caffeine intake time & Timing of last caffeine consumption before sleep & 0.363 & 0.119 \\
Sleeping posture & Quality of body posture during sleep & 0.285 & 0.093 \\
Screen use before sleep & Extent of screen exposure prior to bedtime & 0.259 & 0.085 \\
Consistent sleep schedule & Regularity of sleep schedule over the last month & 0.118 & 0.039 \\
\hline
\end{tabular}
\end{table}

Table~\ref{tab:shap_weights} reports the SHAP-derived weights associated with the actionable variables retained for optimization. Each weight represents the mean absolute contribution of the corresponding variable to the sleep quality prediction model and is used directly as a coefficient in the optimization objective. Larger weights therefore indicate variables whose improvement is expected to yield greater impact on predicted sleep quality. As shown in the table, environmental factors such as room cross-ventilation and nighttime quietness receive relatively high weights, suggesting that modest improvements in the sleep environment may offer substantial benefits. In contrast, behavioral regularity variables, such as sleep schedule consistency, receive smaller weights, reflecting a more moderate marginal influence. These relative differences guide the optimization model in prioritizing interventions while adhering to the minimal-change philosophy introduced in the Problem Definition.

\section{Experimental Setup and Results}
\label{sec: exp_result}
To assess the practical and structural behavior of the proposed framework, a comprehensive experimental evaluation is conducted. The analysis investigates sensitivity to the behavioral resistance parameter, characterizes the benefit–cost trade-off, and demonstrates individualized intervention outcomes. The results provide empirical evidence of the model’s stability, sparsity properties, and decision-support capability.

\subsection{Experimental Design}


\subsubsection{Per-Student Optimization Setup}

The optimization model is solved independently for each student, treating the student’s observed sleep-related profile as a fixed baseline. For a given student, all non-actionable attributes are held constant, while actionable variables are allowed to change subject to feasibility and behavioral constraints. This per-student formulation enables fully personalized recommendations and avoids aggregating heterogeneous behavioral patterns into a single population-level decision. For each individual, the optimization determines whether adjusting any actionable variable by at most one ordinal level is warranted, given the trade-off between expected improvement in sleep quality and the cost of behavioral change. Importantly, the model allows for the possibility that no intervention is recommended when the predicted benefit does not justify disruption, reinforcing its role as a decision-support tool rather than a prescriptive mandate.

\subsubsection{Solver and Implementation Details}

All optimization problems are formulated as mixed-integer linear programs and solved using the open-source \textit{Coin-or Branch and Cut (CBC)} solver through the \textit{PuLP} Python modeling interface \citep{forrest2005cbc, mitchell2011pulp}. CBC is a widely used branch-and-cut solver for MILP problems and is well suited to the scale of the proposed formulation, which involves a small number of integer and binary decision variables per student. Given the limited problem size, all instances are solved to optimality within negligible computational time, ensuring that solver performance does not influence the results. The use of an open-source solver enhances reproducibility and allows the proposed framework to be readily deployed without reliance on proprietary software.

\subsubsection{Behavioral Resistance Parameter Settings}

The optimization objective includes a behavioral resistance parameter \( \lambda \), which penalizes the number of activated interventions and governs the trade-off between predicted sleep quality improvement and lifestyle disruption. To examine the robustness of the proposed framework and to understand how recommendations vary with willingness to change, a sensitivity analysis is conducted over multiple values of \( \lambda \).

Specifically, three representative values are considered, corresponding to different levels of behavioral resistance. These values are summarized in Table~\ref{tab:lambda_values}.

\begin{table}[t]
\centering
\caption{Behavioral Resistance Parameter Settings}
\label{tab:lambda_values}
\begin{tabular}{c l}
\hline
\textbf{\( \lambda \) Value} & \textbf{Interpretation} \\
\hline
0.1 & Low resistance to behavioral change \\
0.2 & Moderate resistance to behavioral change \\
0.3 & High resistance to behavioral change \\
\hline
\end{tabular}
\end{table}

The value \( \lambda = 0.2 \) is used as the primary setting for reporting personalized recommendations, as it reflects a balanced and realistic trade-off between intervention effectiveness and behavioral burden. The remaining values are used to assess sensitivity and to illustrate how the model transitions from proactive to conservative intervention strategies as resistance increases.

\subsection{Sensitivity and Pareto Trade-off Analysis}

Building upon the parameter settings introduced above, the empirical impact of varying \( \lambda \) is examined to characterize the resulting trade-off between intervention intensity and expected benefit.

\begin{figure}[htbp]
\centering
\includegraphics[width=\linewidth]{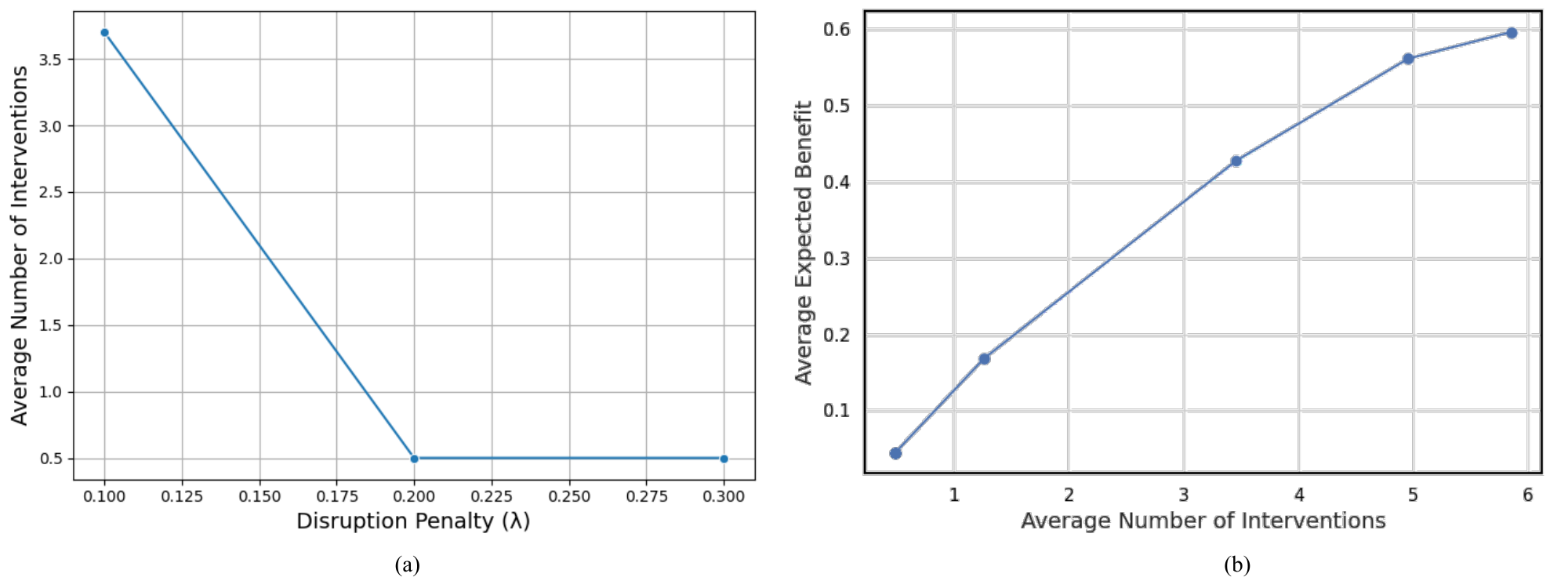}
\caption{Sensitivity and Pareto analysis of the optimization model. (a) Average number of recommended interventions decreases as the behavioral resistance parameter ($\lambda$) increases. (b) Pareto frontier between average expected benefit and intervention count, showing diminishing marginal gains as more interventions are applied.}
\label{fig:lambda_sensitivity}
\end{figure}

Figure~\ref{fig:lambda_sensitivity}(a) presents the average number of recommended interventions per student across different values of \( \lambda \). When \( \lambda = 0.1 \), corresponding to low resistance to behavioral change, the model recommends multiple interventions on average, reflecting a setting in which predicted improvement dominates the disruption penalty. As \( \lambda \) increases to 0.2, the average intervention count decreases sharply, indicating that the optimizer begins to retain only those adjustments with sufficiently high expected marginal benefit. For \( \lambda = 0.3 \), the model becomes markedly conservative, frequently recommending few or no modifications unless the anticipated improvement clearly justifies the associated behavioral cost. This monotonic decline confirms that the framework responds in a controlled and interpretable manner to increasing resistance, producing progressively sparser and more selective intervention sets.

While panel (a) illustrates how intervention intensity varies with \( \lambda \), it does not explicitly quantify the benefit–cost structure underlying this behavior. To address this, Figure~\ref{fig:lambda_sensitivity}(b) presents the Pareto frontier between the average expected improvement (measured as the weighted sum \( \sum_i w_i \Delta x_i \)) and the average number of recommended interventions. The resulting curve is strictly increasing and concave, demonstrating a clear trade-off: larger intervention sets yield greater predicted improvement, but with diminishing marginal returns. In particular, the initial interventions contribute substantial gains in expected benefit, whereas additional modifications provide progressively smaller incremental improvements.

The concave shape of the Pareto frontier provides empirical support for the minimal-change philosophy embedded in the model. Rather than suggesting comprehensive lifestyle overhauls, the optimization identifies a region of efficient trade-off in which modest, targeted adjustments achieve a substantial portion of the attainable benefit. The operating point \( \lambda = 0.2 \) lies near this efficient region, balancing predicted improvement against behavioral burden without encouraging excessive intervention intensity.

\subsection{Ablation Studies}

To assess the structural contribution of key modeling components, two targeted ablation studies are conducted. The objective is to examine the role of the behavioral resistance term and the SHAP-derived weighting mechanism in shaping intervention sparsity and benefit–cost trade-offs. Each ablation modifies one component of the objective function while preserving all feasibility constraints, enabling isolated evaluation of its influence on solution structure.

\paragraph{(1) Removal of Behavioral Resistance Penalty (\(\lambda = 0\)).}
In the first ablation, the disruption penalty term \(\lambda \sum_{i \in \mathcal{A}} z_i\) is removed from the objective, yielding:

\begin{equation}
\max \sum_{i \in \mathcal{A}} w_i \Delta x_i
\label{eq:ablation_no_penalty}
\end{equation}
This formulation maximizes predicted benefit without penalizing the number of activated interventions, effectively eliminating the minimal-change mechanism embedded in the full model.

\paragraph{(2) Equal-Weight Objective (No SHAP Weighting).}
In the second ablation, all importance weights are set to unity, resulting in:
\begin{equation}
\max \sum_{i \in \mathcal{A}} \Delta x_i 
- \lambda \sum_{i \in \mathcal{A}} z_i
\label{eq:ablation_equal_weights}
\end{equation}
This modification removes data-driven prioritization and treats all actionable variables as equally influential, thereby isolating the effect of SHAP-based importance weighting on intervention selection.

\begin{table}[htbp]
\centering
\caption{Ablation study comparison at \(\lambda = 0.2\)}
\begin{tabular}{lcc}
\hline
\textbf{Model Variant} & \textbf{Avg. Interventions} & \textbf{Avg. Benefit} \\
\hline
Full Model & 0.53 & 0.048 \\
No Penalty (\(\lambda = 0\)) & 5.83 & 0.596 \\
Equal Weights & 5.83 & 5.830 \\
\hline
\end{tabular}
\label{tab:ablation}
\end{table}

Table~\ref{tab:ablation} highlights substantial structural differences across model variants. Removing the behavioral resistance term results in a dramatic increase in intervention intensity, with the average number of activated variables rising from 0.53 to 5.83. This confirms that the penalty component is the primary driver of sparsity in the full formulation and is essential for enforcing the minimal-change philosophy. Similarly, replacing SHAP-derived weights with uniform coefficients produces the same intervention count (5.83), indicating that without importance differentiation, the model is incentivized to activate nearly all feasible variables. The resulting benefit value increases mechanically due to the objective structure, but this improvement lacks prioritization and interpretability. Together, these findings demonstrate that both components of the objective function are structurally necessary. The behavioral penalty ensures intervention restraint, while SHAP-based weighting enables selective prioritization of high-impact variables. Their joint inclusion produces sparse, interpretable, and behaviorally realistic recommendations that neither component alone can achieve.

\subsection{Personalized Recommendations}

The personalized intervention results in Table~\ref{tab:student_recommendations} provide several important insights into the behavior of the proposed optimization framework and its ability to translate predictive importance into actionable, individualized guidance. Rather than uniformly recommending changes across all actionable variables, the model selectively identifies interventions whose expected benefit justifies behavioral disruption under the chosen resistance level \( \lambda = 0.2 \).
\begin{table}[htbp]
\centering
\caption{Personalized intervention recommendations for representative students (\(\lambda = 0.2\)).}
\label{tab:student_recommendations}
\begin{tabular}{l c c c}
\hline
\textbf{Actionable Variable} & \textbf{Baseline} & \textbf{Change} & \textbf{Optimized} \\
\hline
\multicolumn{4}{l}{\textbf{Student 874}} \\
Screen use before sleep & 0 & +1 & 1 \\
Consistent sleep schedule & 3 & 0 & 3 \\
Last caffeine intake time & 4 & 0 & 4 \\
Lighting condition & 4 & 0 & 4 \\
Room cross-ventilation & 5 & 0 & 5 \\
Nighttime quietness & 5 & 0 & 5 \\
Sleeping posture & 1 & 0 & 1 \\
\hline
\multicolumn{4}{l}{\textbf{Student 394}} \\
Consistent sleep schedule & 3 & 0 & 3 \\
Last caffeine intake time & 1 & 0 & 1 \\
Lighting condition & 5 & 0 & 5 \\
Screen use before sleep & 4 & 0 & 4 \\
Room cross-ventilation & 3 & 0 & 3 \\
Nighttime quietness & 5 & 0 & 5 \\
Sleeping posture & 3 & 0 & 3 \\
\hline
\multicolumn{4}{l}{\textbf{Student 1220}} \\
Screen use before sleep & 0 & +1 & 1 \\
Sleeping posture & 0 & +1 & 1 \\
Consistent sleep schedule & 4 & 0 & 4 \\
Last caffeine intake time & 4 & 0 & 4 \\
Lighting condition & 5 & 0 & 5 \\
Room cross-ventilation & 3 & 0 & 3 \\
Nighttime quietness & 2 & 0 & 2 \\
\hline
\end{tabular}
\end{table}

For Student~874, the optimization recommends a single intervention involving screen use before sleep. In real terms, this corresponds to shifting from screen exposure in the immediate pre-sleep period (e.g., within the last few minutes before bedtime) to an earlier cutoff, such as discontinuing screen use approximately 30 minutes before sleep. All other behavioral and environmental variables are left unchanged. This outcome suggests that, for this individual, late-night screen exposure represents a dominant and addressable contributor to poor sleep quality, while additional modifications are unlikely to yield proportional gains. From a decision-support perspective, this recommendation is particularly valuable, as it isolates a single, low-effort adjustment with a favorable benefit-to-disruption ratio.

In contrast, the optimization model recommends no interventions for Student~394. Despite moderate baseline values across several actionable variables, the expected improvement associated with modifying any single factor does not outweigh the corresponding behavioral penalty. This result is not a limitation of the framework but rather a key strength: the model explicitly allows for the possibility that maintaining the current state is optimal. Such outcomes highlight the importance of avoiding unnecessary or low-impact recommendations and demonstrate that the framework does not force interventions when empirical evidence does not support them.

For Student~1220, the model identifies two complementary interventions. The first recommends a reduction in screen exposure before bedtime by one ordinal level, corresponding to an earlier cessation of screen use. The second recommends an improvement in sleeping posture, such as transitioning from an ergonomically unfavorable position to a more supportive posture. Notably, no changes are recommended for caffeine intake timing or environmental conditions, indicating that these factors already lie within acceptable ranges for this student. This combination of recommendations illustrates how the framework can suggest multiple, targeted adjustments when their joint expected benefit justifies modest behavioral effort.

Across all three cases, the magnitude of recommended changes is deliberately restrained, with all interventions limited to single-level ordinal adjustments. This pattern reflects the minimal-change philosophy embedded in the formulation and ensures that recommendations remain behaviorally realistic and interpretable. Importantly, the heterogeneity in outcomes from no intervention to one or two targeted adjustments demonstrates the framework’s ability to adapt recommendations to individual profiles rather than relying on population-level heuristics. Taken together, these results underscore the value of integrating explainable machine learning with optimization. SHAP-derived importance weights guide the optimization toward empirically supported interventions, while the MILP formulation balances predicted benefit against behavioral cost. The resulting recommendations are personalized, selective, and grounded in data, offering a practical pathway for decision-support systems aimed at improving sleep quality without imposing excessive lifestyle disruption.

\section{Discussion}
\label{sec: discussion}
Integrating predictive modeling with constrained optimization introduces several conceptual and practical considerations. Although the empirical results clarify the operational behavior of the framework, their broader significance lies in the translation of prediction-driven importance measures into structured decision rules under explicit behavioral constraints. Observed trade-offs between expected benefit and intervention intensity reflect the inherent tension between improvement and feasibility, a balance central to real-world lifestyle interventions. These dimensions merit further discussion from interpretive, practical, and methodological perspectives.

\subsection{Interpretation of Minimal-Change Recommendations}

The recommendations produced by the proposed optimization framework should be interpreted as targeted adjustments that are expected to yield meaningful improvements in predicted sleep quality while minimizing disruption to an individual’s existing routines. Unlike prescriptive approaches that seek to enforce idealized behavioral patterns, the model is explicitly designed to recommend changes only when they are justified by sufficient expected benefit. As a result, solutions in which few or no variables are modified are not only admissible but often optimal, reflecting the underlying trade-off structure of the formulation. This behavior arises directly from the objective function, which balances the expected improvement in sleep quality against a penalty associated with behavioral change. Each potential intervention contributes positively through its SHAP-derived importance weight, while the introduction of binary decision variables imposes a cost for modifying any actionable factor. Consequently, the optimization favors sparse intervention sets, prioritizing changes with the highest marginal impact and suppressing low-impact or redundant adjustments. When the predicted benefit of modifying a variable does not outweigh the associated disruption cost, the model rationally opts to preserve the baseline behavior. The parameter \( \lambda \) plays a central role in controlling the conservativeness of the recommendations. Larger values of \( \lambda \) increase the penalty for behavioral changes, resulting in more conservative solutions that recommend fewer interventions, whereas smaller values allow a greater number of changes when the cumulative expected benefit is sufficiently high. This mechanism provides a transparent and tunable means of modeling behavioral resistance, enabling the framework to adapt to different assumptions about individual willingness to change. Overall, the minimal-change philosophy embedded in the formulation ensures that recommendations are interpretable, selective, and behaviorally realistic, aligning the optimization outputs with practical decision-making rather than idealized outcomes.

\subsection{Practical and Behavioral Implications}

Practical applicability of the proposed framework lies in its focus on everyday sleep-related behaviors rather than clinical diagnosis or treatment. By targeting modifiable lifestyle and environmental factors, the recommendations are framed as supportive guidance that students can realistically adopt within the constraints of academic life. This non-clinical positioning is particularly important in university settings, where sleep-related challenges are widespread but formal medical intervention may be inaccessible or unnecessary for many individuals \citep{becker2018sleep}. Feasibility and acceptability are central to the design of the recommendation mechanism. Interventions are deliberately restricted to small, ordinal adjustments, which reduces cognitive and behavioral burden and increases the likelihood of sustained adherence. Instead of advocating comprehensive lifestyle overhauls, the framework prioritizes incremental changes that align with existing routines. This approach reflects evidence from behavioral research suggesting that modest, targeted adjustments are more likely to be implemented and maintained over time \citep{conn2016medication}. Behavioral resistance is explicitly modeled within the optimization process through a disruption penalty, acknowledging that individuals vary in their willingness and capacity to change. By internalizing resistance rather than treating it as an external constraint, the model naturally limits the number of recommended interventions and avoids overwhelming users with excessive guidance. As a result, recommendation fatigue is mitigated, and the output remains selective and interpretable. From an ethical standpoint, the conservative nature of the recommendations represents a deliberate design choice. Avoiding over-prescription and unrealistic lifestyle demands helps preserve individual autonomy and reduces the risk of unintended negative effects associated with excessive self-monitoring or behavioral pressure \citep{rams2026digital}. Recommendations are generated only when supported by empirical evidence and justified by a favorable trade-off between expected benefit and disruption, ensuring that guidance remains proportionate, transparent, and behaviorally grounded.

\subsection{Limitations}

Several limitations of the proposed framework should be acknowledged. First, the analysis is based on self-reported survey data, which may be subject to recall bias and subjective perception, particularly for behavioral and environmental variables. While standardized instruments such as PSQI were used to mitigate this issue, objective sleep measurements were not available. Second, the optimization formulation relies on ordinal representations of behavioral changes and restricts interventions to single-step adjustments. Although this design choice supports interpretability and feasibility, it simplifies the underlying behavioral dynamics and does not capture more gradual or continuous changes. Third, SHAP-derived importance weights are computed from a population-level predictive model and treated as fixed parameters in the optimization stage. As a result, individual-level heterogeneity in feature effects may not be fully reflected in the optimization coefficients. Finally, the current framework evaluates recommendations based on predicted improvements rather than observed post-intervention outcomes, as longitudinal or experimental validation data were not available. These limitations suggest opportunities for refinement while preserving the core structure and interpretability of the proposed approach.

\subsection{Future Research Directions}

Building on the present findings, several directions for future research warrant further exploration. One natural extension involves personalizing the behavioral resistance parameter \( \lambda \) to better reflect individual differences in willingness or capacity to change. Rather than treating \( \lambda \) as a fixed global parameter, future work could infer personalized values based on historical behavior, preference elicitation, or survey-based measures of readiness for change. This would allow the optimization framework to adapt more closely to individual decision-making profiles. Future research may also explore dynamic or multi-period formulations that capture temporal dependencies in sleep behavior. Extending the current static optimization model to a longitudinal setting would enable the study of sequential interventions, habit formation, and delayed effects of behavioral changes. Such formulations could incorporate feedback from observed outcomes, allowing recommendations to evolve over time as individuals respond to prior interventions.

From a modeling perspective, integrating alternative sources of data represents another promising direction. Combining self-reported survey responses with objective measurements from wearable devices or sleep-tracking applications could improve both predictive accuracy and the calibration of optimization parameters. Additionally, the framework could be extended to accommodate uncertainty in model estimates, for example through robust or stochastic optimization approaches that account for variability in predicted outcomes. Finally, broader empirical validation of the proposed framework through pilot studies or controlled interventions would strengthen its practical relevance. Evaluating how individuals respond to the recommended changes, and whether predicted improvements translate into realized sleep quality gains, would provide valuable insights into the real-world effectiveness of optimization-based decision support for lifestyle interventions.

\section{Conclusion} \label{sec: con}
This study presented a unified predictive–prescriptive framework for personalized sleep quality improvement among university students by integrating machine learning, explainable AI, and mixed-integer linear optimization. Using survey data collected from 418 students and subsequently augmented to 1,339 samples, sleep quality was assessed using the Pittsburgh Sleep Quality Index (PSQI) and modeled based on a diverse set of behavioral, environmental, and lifestyle factors. Benchmarking across multiple machine learning models demonstrated that XGBoost achieved the best predictive performance, attaining a test F1-score of 0.954, along with strong precision and recall, making it well suited for downstream decision support. To move beyond prediction, SHAP-based explainability was employed to quantify the relative influence of actionable variables on sleep quality outcomes. These SHAP-derived importance measures were then embedded into a mixed-integer linear programming formulation that recommends minimal, feasible behavioral adjustments for individual students. The optimization model explicitly balances expected improvement in predicted sleep quality against behavioral disruption, controlled through a penalty parameter. Experimental results show that the model frequently recommends sparse intervention sets, and in many cases no changes at all, reflecting a rational preference for preserving baseline behavior when marginal benefits are insufficient. Sensitivity analysis further demonstrated that increasing the penalty parameter leads to more conservative recommendations, providing a transparent mechanism for modeling behavioral resistance. Ablation analysis confirmed the structural necessity of both the behavioral resistance term and the SHAP-derived weighting mechanism: removing the penalty resulted in near-universal intervention activation, while eliminating importance weighting led to indiscriminate variable selection without meaningful prioritization. These findings reinforce that sparsity and interpretability emerge from the joint interaction of the penalty and data-driven weighting components rather than from the constraints alone. The proposed framework contributes a practical and interpretable approach to personalized decision support in sleep health, translating complex predictive insights into actionable recommendations without imposing unrealistic lifestyle demands. By prioritizing minimal-change interventions grounded in empirical model explanations, the approach addresses a critical gap in existing sleep studies that stop at risk identification without offering decision guidance. Future research may extend this work by incorporating longitudinal or experimental validation, personalizing behavioral resistance parameters, and integrating objective sleep measurements from wearable devices. Such extensions would further strengthen the framework’s applicability and deepen its potential impact on data-driven health decision support.

\section*{Acknowledgments}
We are grateful to Computational Intelligence and Operations Laboratory (CIOL) for all kinds of support and guidance in the work.

\section*{Declarations}
\begin{itemize}
\item Funding : No funding was received to assist with the preparation of this manuscript.
\item Conflict of interest: All authors certify that they have no affiliations with or involvement in any organization or entity with any financial interest or non-financial interest in the subject matter or materials discussed in this manuscript.
\item Ethical Concerns: Not Applicable.
\item Ethics approval and consent to participate: Not Applicable.
\item  Potential Risks and Response: Not Applicable.
\item Use of Generative AI and AI-assisted Technologies: During the preparation of this work the author(s) used ChatGPT in order to reduce grammatical errors and writing clarity. After using this tool/service, the author(s) reviewed and edited the content as needed and take(s) full responsibility for the content of the published article.
\item Data availability: The data used in this study will be made available upon request.
\item Materials availability: The materials underlying this study will be made accessible upon request. 
\item Code availability: The code used in this study will be made available upon acceptance.
\end{itemize}

%% file: sections/appendix.tex
\newpage
\appendix

\section{Nomenclature}
\begin{table}[h]
\centering
\caption{List of symbols and notation used in this article}
\begin{tabular}{ll}
\hline
\textbf{Symbol} & \textbf{Description} \\
\hline
$i$ & Index of actionable variables \\
$n$ & Index of students \\
$\mathcal{A}$ & Set of actionable variables \\
$N$ & Number of samples used for SHAP analysis \\

$x_i^0$ & Baseline value of actionable variable $i$ \\
$\Delta x_i$ & Decision variable representing change applied to $i$ \\
$z_i$ & Binary variable indicating whether intervention $i$ is activated \\

$w_i$ & SHAP-derived importance weight for variable $i$ \\
$\tilde{w}_i$ & Normalized importance weight of variable $i$ \\
$\phi_i^{(n)}$ & SHAP value of variable $i$ for student $n$ \\

$\lambda$ & Behavioral resistance (disruption) penalty parameter \\
$f(\cdot)$ & Trained sleep quality prediction model \\

$\underline{x}_i, \overline{x}_i$ & Lower and upper feasible bounds of variable $i$ \\
$\mathcal{L}$ & Optimization objective function value \\

$\mathcal{F}_1$ & F1-score of classification model \\
$\text{Acc}$ & Classification accuracy \\
$\text{Prec}$ & Precision \\
$\text{Rec}$ & Recall \\

$\eta$ & Learning rate in XGBoost \\
$T$ & Number of boosting trees (estimators) \\
$d$ & Maximum tree depth \\

\hline
\end{tabular}
\end{table}

\section{Sleep Quality Survey Instrument}
All responses refer to the previous one-month period unless otherwise specified.

\subsection{Informed Consent}

\textbf{Consent Statement:} 
“I consent to sharing my sleep data and related information for research purposes.”

Response: (Yes / No)

\subsection*{Demographic Information}
\begin{enumerate}
\item \textbf{Gender:} (Male / Female / Others)
\item \textbf{Age (years):} \underline{\hspace{2cm}}
\item \textbf{Weight (kg):} \underline{\hspace{2cm}}
\item \textbf{Height (feet and inches):} \underline{\hspace{3cm}}
\item \textbf{Profession:} (Student / Employed / Unemployed / Entrepreneur / Other)
\item \textbf{Relationship Status:} (Single / In a Relationship (Not Married) / Breakup/Separated / Married / Divorced/Widowed / Prefer Not to Say)
\item \textbf{What type of area do you live in?} (Urban / Sub-Urban / Rural)
\item \textbf{Where did you live in the past 1 month?} (With family at home / By myself at home / In a shared accommodation/mess / In a university hall)
\item \textbf{At which Floor do you live in?} \underline{\hspace{2cm}}
\item \textbf{Where did you sleep in the past month?} (Floor Bedding / Bed)
\end{enumerate}

\subsection*{Habitual Data}

\begin{enumerate}
\setcounter{enumi}{11}

\item \textbf{What time after evening do you typically consume your last caffeinated drink (coffee, tea, energy drinks)?} (6-8 PM / 8 PM - 10 PM / 10 PM - 12 AM / After 12 AM / I don't drink in this period)

\item \textbf{At average, how many caffeinated drinks (coffee, tea, energy drinks) do you consume per day after evening?} (None / 1 / 2 / 3 / More than 3)

\item \textbf{How long before bed do you have caffeinated drinks (coffee, tea, energy drinks)?} (0-5 minutes before bed / 30 minutes before bed / 1 hour before bed / 2 hour before bed / More than 3 hour before bed)

\item \textbf{How many hours per day do you spend looking at screens (phone, computer, TV)?} (Less than 1 hours / 1-2 hours / 2-4 hours / 4-6 hours / More than 6 hours)

\item \textbf{How long before bed do you stop using screens?} (0-5 minutes before bed / 30 minutes before bed / 1 hour before bed / 2 hour before bed / More than 3 hour before bed)

\item \textbf{How long before bed do you stop reading books?} (0-5 minutes before bed / 30 minutes before bed / 1 hour before bed / 2 hour before bed / More than 3 hour before bed / I don't read books usually)

\item \textbf{What is your usual posture when using a screen before sleeping?} (Sitting upright / Lying on your back / Lying on your stomach / Lying on your side / Reclining (partially lying down) / Standing or walking)

\item \textbf{Which sleeping posture do you prefer most?} (Back sleeping (lying on your back) / Stomach sleeping (lying on your stomach) / Left facing / Right facing / Combination (switching between different postures))

\item \textbf{What activity do you most commonly engage in on your screen before going to sleep?} (Watching videos or movies / Listening to music / Listening to podcasts or audiobooks / Browsing social media or the internet / I don't use screen before sleeping)

\item \textbf{Do you use nicotine products (e.g., cigarettes, vaping)?} (I don't consume in any form / Indirect smoking only (from other people, often) / Indirect smoking only (from other people, regularly) / Rarely / Once a week / Chain smoker (1-2 times a day) / Chain smoker (3+ times a day))

\item \textbf{How much do you agree with the statement "I like to take heavy meals before sleeping"?} (1 - Strongly Disagree / 2 / 3 / 4 / 5 - Strongly Agree)

\item \textbf{How much do you agree with the statement "I was consistent in my sleep schedule in last month (same bedtime and wake-up time daily)"?} (1 - Never / 2 / 3 / 4 / 5 - Always)

\item \textbf{How much clothing do you typically wear while sleeping?} (Heavy Clothing (Full Body Coverage and Extra Heavy Clothings) / Full Clothing (Full Body Coverage) / Partial Clothing (Half Body Cover) / Minimal Clothing (Underwears Only) / No Clothing)

\end{enumerate}

\subsection*{Environmental Data}

\begin{enumerate}
\setcounter{enumi}{24}

\item \textbf{How much do you agree with the statement "I think my bed is very comfortable for sleeping"?} (1 - Strongly Disagree / 2 / 3 / 4 / 5 - Strongly Agree)

\item \textbf{How much do you agree with the statement "I believe lighting affects the quality of my sleep."?} (1 - Strongly Disagree / 2 / 3 / 4 / 5 - Strongly agree)

\item \textbf{How much do you agree with the statement "I believe my sleeping environment is quiet."?} (1 - Strongly Disagree / 2 / 3 / 4 / 5 - Strongly agree)

\item \textbf{How much do you agree with the statement "I believe there were adequate cross-ventilation (airflow from both sides) in my room."?} (1 - Strongly Disagree / 2 / 3 / 4 / 5 - Strongly agree)

\end{enumerate}

\subsection*{Physiological Data}

\begin{enumerate}
\setcounter{enumi}{28}

\item \textbf{On average, how much physical activity (exercise, walking, etc.) did you do daily in the previous month?} (Less than 15 minutes / 15-30 minutes / 30-60 minutes / 60-120 minutes / More than 120 minutes)

\item \textbf{How many hours per day do you spend working or studying?} (Less than 1 hours / 1-2 hours / 2-4 hours / 4-6 hours / More than 6 hours)

\item \textbf{Are you a student?} (Yes / No)

\end{enumerate}

\subsection*{Psychological Data}

\begin{enumerate}
\setcounter{enumi}{31}

\item \textbf{Did you have any exams last month?} (Yes / No)

\item \textbf{Did you experience any changes in your sleep patterns during the time of your exams last month?} (Yes / No)

\item \textbf{Did stress from exams affect the quality of your sleep last month?} (Yes / No)

\setcounter{enumi}{34}

\item \textbf{How much do you agree or disagree with the statement "During the past month, I often felt stress because of my financial situation."?} (1 - Strongly Disagree / 2 / 3 / 4 / 5 - Strongly agree)

\item \textbf{How much do you agree or disagree with the statement "In the past month, I had headache or neck pain frequently."?} (1 - Strongly Disagree / 2 / 3 / 4 / 5 - Strongly agree)

\end{enumerate}

\subsection*{Pittsburgh Sleep Quality Index (PSQI)}

\begin{enumerate}
\setcounter{enumi}{36}

\item \textbf{During the past month, what time have you usually gone to bed at night?} \underline{\hspace{4cm}}

\item \textbf{During the past month, how long (in minutes) has it usually taken you to fall asleep each night?} \underline{\hspace{3cm}}

\item \textbf{During the past month, what time have you usually gotten up in the morning?} \underline{\hspace{4cm}}

\item \textbf{During the past month, how many hours of actual sleep did you get at night? (This may be different than the number of hours you spent in bed.)} \underline{\hspace{3cm}}

\item \textbf{During the past month, how often have you had trouble sleeping because you...}

\begin{enumerate}
\item[a)] Cannot get to sleep within 30 minutes (Not during the past month / Less than once a week / Once or twice a week / Three or more times a week)

\item[b)] Wake up in the middle of the night or early morning (Not during the past month / Less than once a week / Once or twice a week / Three or more times a week)

\item[c)] Have to get up to use the bathroom (Not during the past month / Less than once a week / Once or twice a week / Three or more times a week)

\item[d)] Cannot breathe comfortably (Not during the past month / Less than once a week / Once or twice a week / Three or more times a week)

\item[e)] Cough or snore loudly (Not during the past month / Less than once a week / Once or twice a week / Three or more times a week)

\item[f)] Feel too cold (Not during the past month / Less than once a week / Once or twice a week / Three or more times a week)

\item[g)] Feel too hot (Not during the past month / Less than once a week / Once or twice a week / Three or more times a week)

\item[h)] Had bad dreams (Not during the past month / Less than once a week / Once or twice a week / Three or more times a week)

\item[i)] Have pain (Not during the past month / Less than once a week / Once or twice a week / Three or more times a week)

\item[j)] Other reason(s), please describe: \underline{\hspace{8cm}}

\end{enumerate}

\item \textbf{During the past month, how would you rate your sleep quality overall?} (Very good / Fairly good / Fairly bad / Very bad)

\item \textbf{During the past month, how often have you taken medicine to help you sleep (prescribed or "over the counter")?} (Not during the past month / Less than once a week / Once or twice a week / Three or more times a week)

\item \textbf{During the past month, how often have you had trouble staying awake while driving, eating meals, or engaging in social activity?} (Not during the past month / Less than once a week / Once or twice a week / Three or more times a week)

\item \textbf{During the past month, how much of a problem has it been for you to keep up enough enthusiasm to get things done?} (No problem at all / Only a very slight problem / Somewhat of a problem / A very big problem)

\end{enumerate}

\subsection*{Confidentiality Statement}

All responses were anonymized. No personally identifiable information was collected. Data were analyzed in aggregate form and used solely for research purposes.